\documentclass[runningheads]{llncs}

\usepackage[T1]{fontenc}
\usepackage{graphicx}
\usepackage{booktabs}
\usepackage{amsmath}
\usepackage{algorithm}
\usepackage{algpseudocode}
\usepackage{url}

\usepackage{enumitem}

\usepackage{xcolor}
\usepackage{hyperref}

\hypersetup{
    colorlinks=true,
    urlcolor=blue,
    citecolor=blue,
    linkcolor=blue
}

\usepackage{fancyhdr}
\begin{document}

\title{A Modular Agent for Reliable and Auditable Spatial Relation Verification in CT Scans}

\titlerunning{Modular Agent for Spatial Relation Verification}

\author{Simon Vincent Abel\inst{1} \and
Heiko Hillenhagen\inst{2} \and
Michael Götz\inst{2} \and
Timo Ropinski\inst{1} \and
Ayhan Can Erdur\inst{3,4,*} \and
Daniel Santak Wolf\inst{1,2,*}
}

\authorrunning{S. Abel et al.}

\institute{Visual Computing Group, Institute of Media Informatics, Ulm University, Germany  \and
Diagnostic and Interventional Radiology, Ulm University Hospital, Germany \and
Chair for AI in Healthcare and Medicine, Technical University of Munich (TUM) and TUM University Hospital, Germany \and 
Department of Radiation Oncology, TUM University Hospital, Germany \\
\textsuperscript{*}\,\emph{Shared Last Authorship}\\[4pt]
\email{daniel.wolf@uni-ulm.de}
}

\maketitle
\thispagestyle{fancy}

\begin{abstract}
Reliable spatial understanding is an important prerequisite for future medical vision-language systems that aim to support radiological report generation and structured image understanding. While modern vision-language models (VLMs) show promising performance on many medical imaging tasks, recent evidence suggests they remain weak in controlled spatial reasoning and often fail to reliably ground spatial relations in image evidence. Given that radiological reasoning hinges on understanding the relative positions of anatomical structures and findings, this spatial weakness poses risks to diagnostic accuracy. 
We present a modular medical imaging agent for binary spatial relation verification in axial CT slices. Instead of directly predicting spatial answers end-to-end, the system decomposes the task into explicit stages: language parsing, anatomical localization, and deterministic geometric verification. Natural-language queries are converted into structured relation tuples, queried organs are localized with a YOLO-based detector, and the final spatial decision is computed from object centers using deterministic geometric rules.
We evaluate the approach on the held-out MIRP spatial QA benchmark and compare it against representative end-to-end VLM baselines. The best-performing hybrid configuration reaches 94.1\% accuracy and 94.2\% F1, outperforming direct Qwen2-VL prompting by 42.5 percentage points in accuracy, while preserving interpretable intermediate representations and auditable reasoning stages. The results suggest that explicit modular spatial verification can serve as a promising building block for future report-oriented medical imaging agents. We release our code to support reproducibility and further research at \url{https://github.com/DeveloperNomis/MICCAI-Medical-Agent}.

\end{abstract}

\section{Introduction}

Vision-Language Models (VLMs) are increasingly explored for radiological applications such as conversational assistance, image interpretation, and report generation~\cite{pellegrini2025radialog,sellergren2025medgemma,wang2024qwen2}. A useful radiology assistant should not only generate plausible descriptions, but also make spatial statements traceable to verifiable image evidence. In such settings, spatial understanding is not a secondary capability. Radiology reports are expected to describe findings with clear anatomical localization~\cite{european2011good}, and spatial language is a central component of radiological reporting~\cite{datta2020understanding}. Failures of spatial localization may have serious clinical consequences, including wrong-level spine surgery or incorrect interpretation of relationships between tumors and adjacent anatomical structures~\cite{agolia2019preventing,kang2021factors}. Thus, medical VLMs intended for report-oriented image understanding must be able to ground spatial statements in verifiable image evidence.
Despite recent progress, spatial reasoning remains a known weakness of current VLMs. Prior work has shown that VLMs struggle with compositional relations and basic spatial concepts such as left/right or above/below~\cite{yuksekgonul2023vision,kamath2023s}. In medical imaging, this limitation is particularly relevant: on controlled spatial reasoning tasks on medical images, representative strong VLMs remain close to chance level~\cite{wolf2025your}. Beyond the possible inaccuracies in the answers, end-to-end VLMs are also limited in providing verifiable evidence for their spatial decisions. Weak visual grounding and hallucination therefore remain central bottlenecks for clinical deployment of medical VLMs~\cite{gu2026medvh,luo2025vividmed}. Without reliable spatial grounding, VLMs cannot be expected to consistently describe anatomical locations and relationships in radiology reports.

Recently, a promising alternative to monolithic prediction has emerged in agentic AI, where intermediate representations are exposed and parts of the reasoning process are delegated to specialized tools. While prior work has structured radiology reports via text-derived entity-relation graphs~\cite{datta2020understanding,jain2021radgraph}, recent agentic medical imaging systems take a multimodal approach, using language models to orchestrate specialized tools across tasks and modalities. Examples include general medical imaging agents like MMedAgent~\cite{li2024mmedagent}, as well as radiology-oriented agents for chest CT and X-ray interpretation~\cite{fallahpour2025medrax,roschewitz2026radagent}, neuro-radiological image analysis~\cite{hoopes2024voxelprompt,erdur2026agentic}, and report generation~\cite{yi2025multimodal}. These works demonstrate the promise of agentic medical AI, but they do not specifically evaluate whether anatomical spatial relations can be verified as controlled, image-grounded, and auditable intermediate facts.
To address this, we propose a modular, tool-using medical imaging agent framework that explicitly decomposes user queries and delegates them to specialized components.  From the user's perspective, the system maintains the familiar conversational interface of a conventional VLM. Internally, however, spatial reasoning is delegated to a dedicated verification tool. Given an axial CT slice and a query such as ``Is the liver left of the spleen?'', the proposed pathway follows a hybrid design: a VLM/LLM component is used for prompt handling, task routing, and language-to-query conversion, while anatomical localization is performed by a dedicated YOLO-based detector. A deterministic geometric module then verifies the queried spatial relation from the detected object centers. No neural model directly predicts the final truth value; instead, the spatial decision is derived from explicit image-space geometry. Finally, while instantiated here for spatial reasoning, this modular architecture acts as a foundation that can be readily extended with additional tools.

We study spatial relation verification as a controlled subproblem of report-oriented radiological image understanding. Isolating this capability provides a rigorous, reproducible setting to demonstrate how a modular agent can solve clinically relevant reasoning more reliably than end-to-end VLMs. Because language parsing, anatomical localization, and geometric verification are exposed as separate stages, our system shifts the paradigm from black-box prediction to a fully auditable process, where failures are explicitly attributed to individual modules. The modular design can also be extended with additional tools for presence verification, measurement-oriented reasoning, or structured report generation, while the present work evaluates the spatial verification component in isolation.

To evaluate our approach, we utilize the MIRP spatial question-answering benchmark~\cite{wolf2025your} presented at MICCAI 2025, comparing the proposed modular agent against representative end-to-end VLM baselines under a unified evaluation protocol. The main contributions of this work are:
\begin{enumerate}
    \item Modular Agent Architecture: We introduce a task-oriented medical imaging agent that maintains a standard conversational VLM interface while internally delegating visual reasoning to specialized, auditable tools.
    \item Explicit Spatial Verification: We design an explicit spatial reasoning pathway that replaces hallucination-prone VLM predictions with a combination of robust anatomical localization and deterministic geometric verification.
    \item Stage-Wise Evaluation and Superior Performance: We demonstrate that the agent substantially outperforms direct end-to-end VLMs (MedGemma, Qwen2-VL) while our stage-wise analysis shifts evaluation from mere black-box accuracy to precise, module-level failure attribution.
\end{enumerate}
Code to extend the agentic framework with new
medical reasoning tools is available at \url{https://github.com/DeveloperNomis/MICCAI-Medical-Agent}.

\section{Method}
We implement spatial relation verification as one pathway within a modular medical imaging agent driven by a core Vision-Language Model (VLM). Our goal is to keep the external interface comparable to a conventional VLM, while making the internal reasoning process explicit and auditable through tool use. Given a CT slice and a user prompt, the agent extracts the spatial question, routes it to a specialized verification pathway, localizes the queried structures, and verifies the relation via deterministic geometry.

\begin{figure*}[t]
\centering
\includegraphics[width=0.86\textwidth]{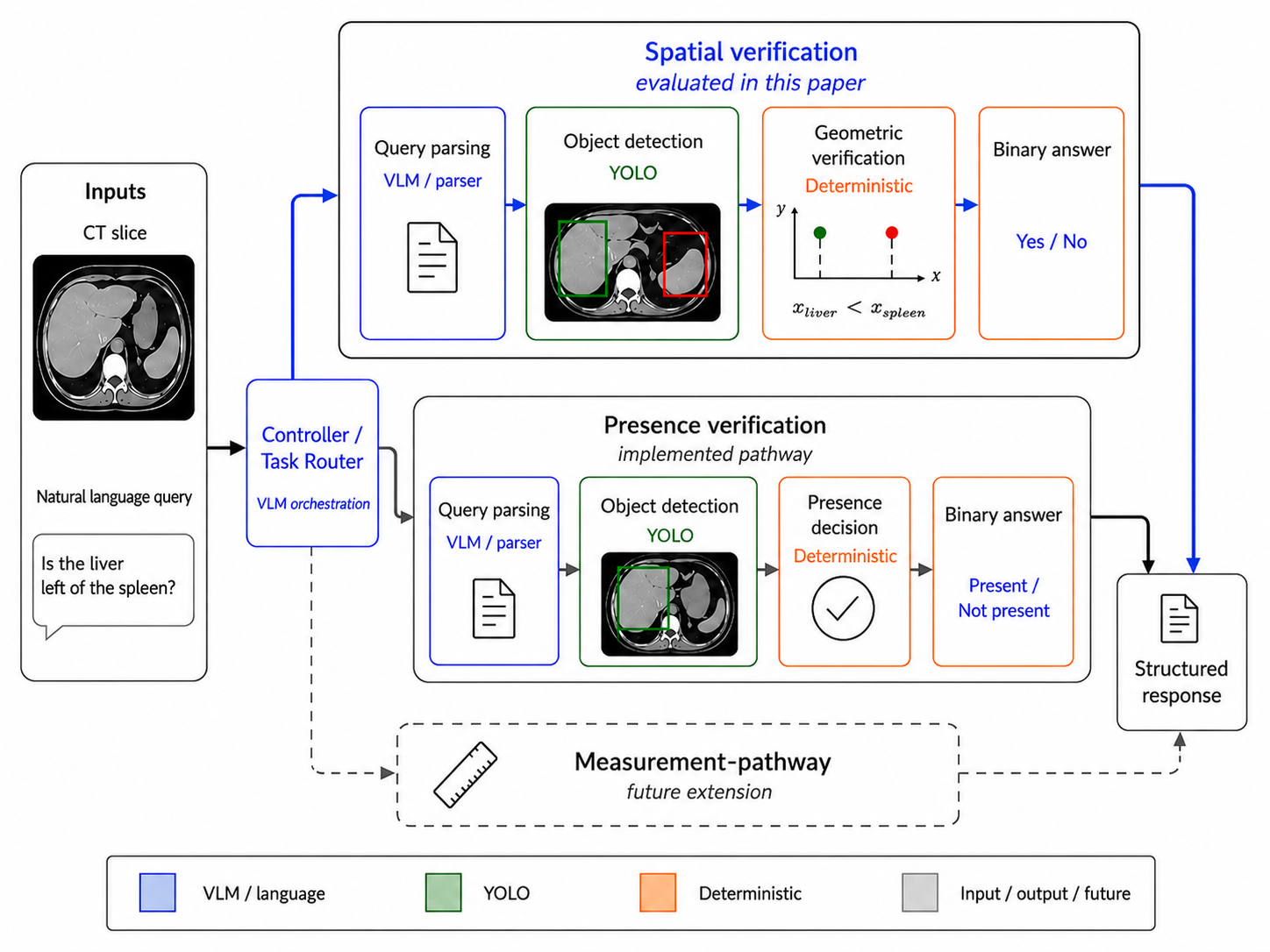}
\caption{\textbf{Hybrid task-oriented agent for medical image reasoning.} To ensure reliable and auditable answers, our agent delegates specific visual reasoning tasks to specialized components. A central VLM-based controller receives a CT slice and a user prompt, extracts the core query, and routes it to an appropriate task-specific pathway. The pathway processes the image and returns a structured, verifiable result. Box colors indicate module type: VLM/language (blue), YOLO perception (green), deterministic logic (orange), and input/output or future extension (gray). Blue arrows mark the evaluated spatial-verification flow; dashed elements mark future extensions. In the evaluated pathway, the binary answer is computed from YOLO detections and deterministic geometry rather than directly predicted by the VLM.}
\label{fig:framework_overview}
\end{figure*}

Figure~\ref{fig:framework_overview} shows the overall framework. A VLM controller receives an image-query pair and selects an appropriate task pathway. We use the LangChain framework~\cite{LangChain} as an orchestration layer for prompt handling, tool invocation, routing, and intermediate state management. LangChain is not used as an additional reasoning model; the evaluated spatial decision is produced by the dedicated spatial verification pathway. The current system supports spatial and presence verification as implemented tools, while measurement-oriented reasoning represents a future modular extension. This work focuses on the spatial pathway, providing a controlled setting for evaluating the benefits of explicitly delegating image-grounded reasoning to specialized modules.

For fair benchmarking, the hybrid agent receives the same external prompts as the direct VLM baselines. However, instead of generating a hallucination-prone direct answer, the controller extracts the core question and routes it to the spatial verification module. This standardized approach preserves a natural user interface while allowing the hybrid agent to apply explicit modular reasoning internally.

\noindent The spatial pathway consists of four connected stages:
\begin{enumerate}[label=\arabic*., leftmargin=*, noitemsep, topsep=0pt, parsep=0pt]
    \item \textbf{Tuple extraction:} The question is converted into a structured tuple (e.g., ``Is the liver left of the spleen?'' $\rightarrow$ \texttt{(liver, left of, spleen)}). This is handled by a lightweight parser with a VLM fallback to robustly map complex natural language into an explicit symbolic representation.
    \item \textbf{Ontology matching:} Extracted entities are mapped to canonical detector classes. If this fails, the agent prevents unsupported queries by immediately returning an invalid result with an audit entry.
    \item \textbf{Localization:} The agent queries a YOLO-based detector to localize the structures in the CT slice, passing only the highest-confidence bounding box per class to the next step.
    \item \textbf{Deterministic verification:} Spatial relations are evaluated from the horizontal or vertical ordering of the detected object centroids. The truth value is derived explicitly from tool-provided coordinates, avoiding neural prediction errors.
\end{enumerate}
Finally, the agent returns a structured task result and records the relevant intermediate states for auditability, including the original prompt, extracted query, ontology match, selected detections, binary result, and audit information.

Crucially, because the intermediate stages are explicit, the system enables fine-grained failure attribution. For any incorrect or invalid case, the earliest failing stage can be identified. This allows errors to be precisely isolated and assigned to failure cases such as question extraction, routing, parsing/query extraction, ontology matching, missing detections, imprecise localization, geometric ambiguity, or formatting and runtime errors.

\section{Experimental Setup}

We compare direct end-to-end VLM prompting against our modular hybrid agent under the same external prompt interface. The framework is model-agnostic; here, we instantiate it with two representative VLMs with publicly available weights: MedGemma~4b~\cite{sellergren2025medgemma} as a medical-domain model and Qwen2-VL~7b~\cite{wang2024qwen2} as a general-purpose model. Each model is evaluated both directly and as part of the hybrid agent. The evaluation is performed on the MIRP RQ1 benchmark \cite{wolf2025your}, which consists of axial CT slices paired with binary questions about pairwise spatial relations between anatomical structures.
We split the MIRP RQ1 benchmark dataset into a training and testing cohort. The resulting held-out test set used for our evaluation contains 938 image-question pairs.
The anatomical detector is trained on the training cohort with segmentation-derived bounding boxes, supplemented by images from the AMOS~\cite{ji2022amos} and BTCV~\cite{landman2015miccai} datasets. We enforce strict patient-level isolation: no patients from the test cohort are present in object detection training. The final detector covers 61 anatomical classes and is trained on 421{,}023 instances. 


In the direct setting, the VLM receives the CT slice and the full prompt, then reports a binary answer. In the hybrid setting, the same prompt is provided to the agent. The agent extracts the intended query from the prompt, routes it to the spatial verification pathway, and returns a binary answer after modular processing. Thus, all systems are evaluated using the same user input, while their internal reasoning architectures differ. Direct VLM decoding was performed deterministically with temperature set to 0.
All evaluated systems receive the following prompt, where \texttt{\{question\}} is replaced by a benchmark question such as ``Is the liver left of the spleen?'':

\begin{quote}
\small
\texttt{This is a 2D axial CT slice.}

\texttt{Question: \{question\}}

\texttt{Answer the question with exactly one character:}

\texttt{1 if the statement is true.}

\texttt{0 if the statement is false.}

\texttt{Do not output any explanation.}
\end{quote}

We report accuracy, F1 score, and invalid-output rate. Strict evaluation counts invalid, failed, or non-binary outputs as incorrect. For the hybrid agent, we additionally perform stage-wise failure attribution. Each incorrect or invalid case is assigned to the earliest stage at which the expected intermediate representation fails. This analysis is used to determine whether remaining errors arise from prompt handling, routing, parsing, ontology matching, detection, localization, geometric ambiguity, or output formatting.

\section{Results}
Table~\ref{tab:main_unified_results} reports the main comparison between our direct and hybrid VLM configurations. To contextualize our performance, the table also includes state-of-the-art results from previously reported VLMs on the current MIRP RQ1 leaderboard \cite{wolf2025your}. The direct VLM baselines remain close to chance level, with accuracies around 52\%, whereas the hybrid-agent variants achieve substantially higher performance. MedGemma + hybrid agent reaches 91.6\% accuracy, and Qwen2-VL + hybrid agent obtains the best result with 94.1\% accuracy and 94.2\% F1. This corresponds to an absolute accuracy gain of 42.5 percentage points over direct Qwen2-VL prompting. No invalid binary outputs were observed.

\begin{table}[ht]
\caption{Main comparison. Prior-work baselines from the current leaderboard of the MIRP Benchmark \cite{wolf2025your} are included for comparison and were not re-evaluated in this work (GPT4o OpenAI \cite{hurst2024gpt}, Gemma~3~12b Google \cite{gemma2025technicalreport}, Pixtral~12b Mistral \cite{agrawal2024pixtral}, MedGemma~4b Google \cite{sellergren2025medgemma}). Because our evaluations were performed on the held-out test set, the evaluated data differs from the leaderboard. Therefore, these are not strictly paired comparisons, which also accounts for the difference between the previously reported and our newly evaluated MedGemma results. Accuracy and F1 are reported in percentages.}
\label{tab:main_unified_results}
\centering
\small

\begin{minipage}{0.43\textwidth}
\centering
\textit{Prior work: MIRP RQ1}\\[2pt]
\begin{tabular}{@{}lcc@{}}
\toprule
\textbf{System} & \textbf{Acc.} & \textbf{F1} \\
\midrule
GPT-4o~\cite{wolf2025your} & 50.5 & 35.2 \\
Gemma~3~\cite{wolf2025your} & 50.9 & 57.3 \\
Pixtral~\cite{wolf2025your} & 50.7 & 43.2 \\
MedGemma~\cite{wolf2025your} & 50.3 & 60.7 \\
\bottomrule
\end{tabular}
\end{minipage}
\hfill
\begin{minipage}{0.53\textwidth}
\centering
\textit{This work: held-out MIRP RQ1 test set}\\[2pt]
\begin{tabular}{@{}lcc@{}}
\toprule
\textbf{System} & \textbf{Acc.} & \textbf{F1} \\
\midrule
MedGemma & 51.8 & 67.2 \\
MedGemma + hybrid & 91.6 & 91.3 \\
Qwen2-VL & 51.6 & 56.8 \\
Qwen2-VL + hybrid & \textbf{94.1} & \textbf{94.2} \\
\bottomrule
\end{tabular}
\end{minipage}

\end{table}

\paragraph{Stage-wise failure attribution}

Beyond the final binary answer, the hybrid agent returns structured intermediate information including the original prompt, extracted question, parsed spatial tuple, ontology-matched class names, selected detections, geometric comparison, final answer, and audit information. This output is used to inspect successful cases and to analyze failures, as summarized in
Table~\ref{tab:failure_attribution}. Each failed case is assigned to the earliest pipeline stage that prevents the correct final answer from being produced.

\begin{table}[t]
\caption{Stage-wise failure attribution for incorrect or invalid hybrid-agent outputs of the best-performing hybrid-agent configuration (Qwen2-VL + hybrid agent). Percentages are computed over the 55 incorrect or invalid cases. Failures were assigned to the earliest identifiable failing stage. In LLM-fallback cases where no separate pre-mapping raw query was stored, incorrect detector-aligned queries were conservatively attributed to parsing/query extraction.}
\label{tab:failure_attribution}
\centering
\small
\begin{tabular}{lccp{0.25\columnwidth}}
\toprule
\textbf{Failure stage} & \textbf{Count} & \textbf{\%} & \textbf{Typical reason} \\
\midrule
Question extraction & 0 & 0.0 & malformed prompt \\
Routing & 0 & 0.0 & wrong pathway \\
Parsing / query extraction & 15 & 27.3 & wrong entity or relation \\
Ontology matching & 0 & 0.0 & detector-class mapping failure \\
Missing detection & 12 & 21.8 & organ not detected \\
Imprecise localization & 28 & 50.9 & center error changes relation \\
Geometry ambiguity & 0 & 0.0 & borderline relation \\
Formatting/runtime & 0 & 0.0 & invalid output or exception \\
\bottomrule
\end{tabular}
\end{table}

The stage-wise analysis shows that most remaining errors of the best performing hybrid configuration arise from perception-related failures. Among the 55 incorrect cases, imprecise localization from the YOLO model is the dominant source with 28 cases (50.9\%), followed by parsing/query-extraction errors with 15 cases (27.3\%) and missing detections with 12 cases (21.8\%). No errors were attributed to question extraction, routing, ontology matching, geometry ambiguity, or formatting/runtime failures under this attribution rule.

\section{Discussion}

The unified prompt protocol directly tests whether explicit modular reasoning provides an advantage over end-to-end VLM prompting when the external interface is held constant. Both direct VLMs and the hybrid agent receive the same CT slice and the same prompt. The difference lies only in how the answer is produced internally.
While the current MIRP RQ1 benchmark baselines show that state-of-the-art VLMs perform at near-chance level, our modular agent effectively solves this benchmark challenge. This demonstrates that explicit spatial verification can successfully bridge the visual grounding gap that conventional models fail to address.
The modular design has two main advantages. First, spatial verification is grounded in explicit anatomical detections and deterministic coordinate comparisons rather than an implicit multimodal prediction. Second, the system exposes intermediate representations that make failures auditable. Because the final geometric verifier is deterministic, cases with correct query extraction, ontology mapping, and selected detections can be traced directly to the corresponding coordinate comparison. This is particularly relevant for medical image reasoning, where a plausible answer is insufficient if it cannot be traced to image evidence.
The failure-attribution analysis is central to this interpretation. By assigning each failed case to the earliest failing stage, the analysis shows where future improvements should focus. In contrast, direct VLM outputs do not expose such intermediate stages, making it difficult to determine whether an error originates from language understanding, visual grounding, or spatial reasoning. At the same time, the analysis shows that the agent's performance is inherently bounded by the quality of its underlying tools: imprecise YOLO localization accounts for 28 of the 55 remaining errors (50.9\%), and missing detections account for another 12 cases (21.8\%).
While the proposed framework demonstrates considerable advantages in accuracy and auditability, the current implementation has several limitations. It evaluates pairwise 2D spatial relations in axial CT slices and does not address full volumetric reasoning, distance-sensitive relations, containment, overlap, or multi-structure consistency. Instead, spatial verification is studied as a controlled intermediate capability that could support future report-oriented medical imaging agents. However, the modular nature of the agent ensures that tools for 3D routing or volumetric measurement can be integrated into the same auditable architecture in future work.

\section{Conclusion}

We propose a modular medical imaging agent for language-grounded spatial relation verification in axial CT slices. The agent uses the same external prompt interface as end-to-end VLM baselines, but internally decomposes the task into question extraction, structured parsing, ontology matching, anatomical localization, and deterministic geometric verification. The proposed design leads to gains in both performance and auditability by replacing end-to-end spatial prediction with explicit, tool-based verification. While standard end-to-end VLMs operate near chance-level accuracy on this spatial reasoning task, the hybrid agent achieves markedly higher accuracy. Furthermore, by exposing intermediate reasoning stages, the agent enables errors to be attributed to specific modules rather than remaining hidden inside an end-to-end multimodal prediction.
Spatial relation verification serves as an initial proof of concept for explicit tool use in medical image reasoning and establishes a foundation for extending the modular agentic framework to additional auditable clinical pathways.

\begin{credits}
\subsubsection{\ackname} This study was funded by the German Federal Ministry of Research, Technology and Space BMFTR as part of the University Medicine Network 3.0 (Project: RACOON, 01KX2524) and by the German Research Foundation DFG (Project: KEMAI, GRK 3012 – 520750254). The authors acknowledge support by the state of Baden-Württemberg through bwHPC.

\subsubsection{\discintname}
The authors have no competing interests to declare that are relevant to the content of this article.
\end{credits}

\bibliographystyle{splncs04}
\bibliography{Paper-29}

\end{document}